\pdfoutput=1
\documentclass[letterpaper, 10pt, conference]{ieeeconf}
\IEEEoverridecommandlockouts
\usepackage{cite}
\usepackage{amssymb}
\usepackage{graphicx}
\usepackage{multirow}
\usepackage{booktabs}
\usepackage{tabularx}
\usepackage{array}
\usepackage[table]{xcolor}
\usepackage{url}
\usepackage{xspace}
\usepackage{textcomp}
\usepackage{caption}
\usepackage{balance}
\usepackage{amsmath,amsfonts,bm}

\def\eqref#1{equation~\ref{#1}}

\def\1{\bm{1}}

\DeclareMathAlphabet{\mathsfit}{\encodingdefault}{\sfdefault}{m}{sl}
\SetMathAlphabet{\mathsfit}{bold}{\encodingdefault}{\sfdefault}{bx}{n}

\usepackage[colorlinks,linkcolor=blue!55!black,citecolor=blue!55!black,urlcolor=blue!55!black]{hyperref}

\DeclareCaptionLabelSeparator{ieeefig}{.~~}
\DeclareRobustCommand{\tabnote}[1]{\\[1pt]{\normalfont\footnotesize #1}}

\newcommand{\cmark}{\ensuremath{\checkmark}}
\newcommand{\xmark}{\ensuremath{\times}}

\definecolor{sectiongray}{HTML}{F2F3F5}
\definecolor{oursblue}{HTML}{EAF2FF}

\newcolumntype{L}{>{\raggedright\arraybackslash}X}
\newcolumntype{C}{>{\centering\arraybackslash}p{0.12\linewidth}}

\makeatletter
\DeclareRobustCommand\onedot{\futurelet\@let@token\@onedot}
\def\@onedot{\ifx\@let@token.\else.\null\fi\xspace}

\makeatother

\title{\LARGE \bf
EgoPush: Egocentric Multi-Object Rearrangement for Mobile Robots
via Constrained Teacher Observability
}

\author{Boyuan An, Zhexiong Wang$^{*}$, Yipeng Wang$^{*}$, Jiaqi Li, Sihang Li$^{\dagger}$, Jing Zhang$^{\dagger}$, and Chen Feng$^{\dagger}$\\[7pt]
New York University%
\thanks{$^{*}$Equal contribution. $^{\dagger}$Corresponding authors: \texttt{\{sl10496, z.jing, cfeng\}@nyu.edu}.}%
}

\IEEEaftertitletext{%
  \vspace{-3mm}\vspace{-12pt}%
  \begin{center}%
    \includegraphics[width=1\textwidth]{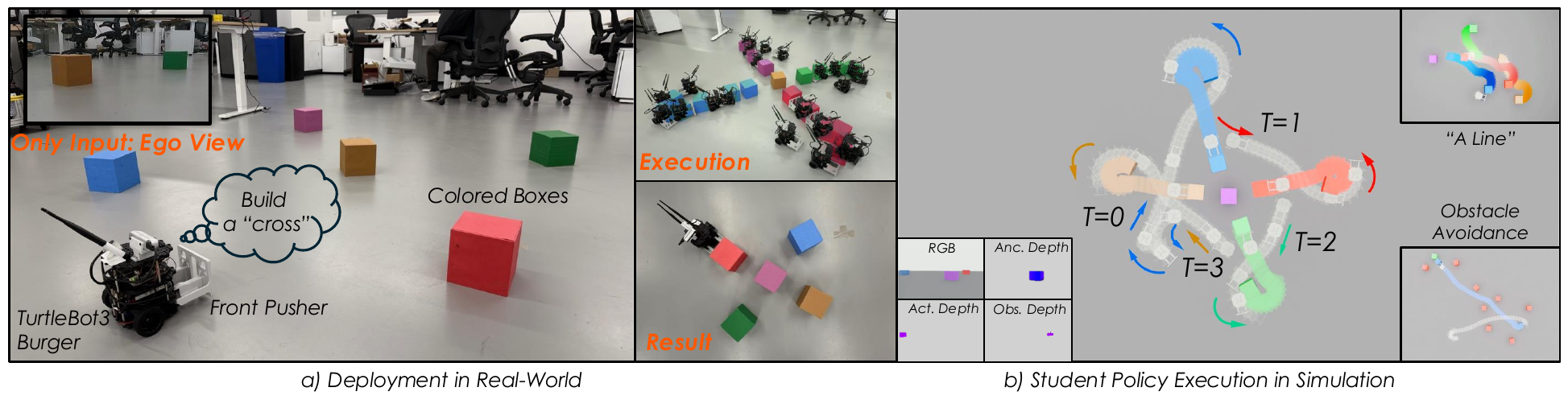}%
  \end{center}%
  \vspace{-3mm}%
  \captionof{figure}{\textbf{EgoPush}: a mobile robot rearranges objects in a controlled real-world arena (left, cross-shape cubes only) and in simulation (right, cross/line formations and cube/cylinder/prism geometries).}%
  \label{fig:head}%
  \vspace{0.6\baselineskip}%
}

\begin{document}
\bstctlcite{IEEEexample:BSTcontrol}
\maketitle
\thispagestyle{empty}
\pagestyle{empty}

\begin{abstract}
Humans rearrange objects in cluttered environments using egocentric perception, actively moving to keep task-relevant spatial cues in view. Mobile robots have not matched this: rearrangement is usually built on a global pose estimate or a map, which is exactly what a robot carrying one camera lacks, while pushing keeps changing the scene it would have to be built from. We present EgoPush, which pushes objects into anchor-relative formations from onboard RGB-D alone, with no global localization, external tracking, or map at deployment, and transfers zero-shot to a TurtleBot in controlled and visually cluttered scenes. What makes this learnable turns out to be a property of the teacher rather than of the student: three privileged teachers trained with identical rewards, architecture, and hyperparameters all exceed $98\%$ success, yet their distilled egocentric students reach $0\%$, $54.8\%$, and $87.3\%$, the only variable being the teacher's observation function. EgoPush therefore trains the teacher under egocentric observability constraints, restricting it to visibility-limited cues and revealing target references only when the anchor is centrally visible, so that its supervision is recoverable by a depth-based student distilled online. Making the teacher trainable in the first place needs two further pieces: a role-grouped object-centric interface shared by teacher and student, and stage-wise temporally decayed rewards for long-horizon credit assignment. Videos, the playable task, and code are available at \href{https://ai4ce.github.io/EgoPush/}{ai4ce.github.io/EgoPush}.
\end{abstract}

\section{Introduction}
\label{sec:intro}

Humans routinely rearrange objects in cluttered environments without precise \textit{global} coordinates: they move around, re-check task-relevant cues, recover from occlusions, and keep an approximate sense of \textit{relative} spatial relations, e.g., keeping the relevant chairs and the table in view while arranging chairs around it.

Enabling similar capabilities on mobile robots remains challenging: global localization, external tracking, and complete object-state estimation are often unavailable or fragile, while sparse landmarks, interaction-induced object motion, and a limited field of view force the robot to coordinate navigation, perception, and contact-rich pushing over a long horizon. This motivates our question: how can a mobile robot rearrange multiple objects from egocentric observations alone, without explicit global state estimation at deployment?

We address this question with EgoPush, a policy-learning framework for an egocentric multi-object rearrangement task.
The robot observes the scene from a single onboard RGB-D camera and sequentially pushes objects into anchor-relative target formations, such as cross- and line-shaped layouts (Fig.~\ref{fig:head}).
We intentionally use simplified geometries and canonical formations to isolate the core difficulties of egocentric rearrangement.

Prior work on non-prehensile manipulation assumes access to precisely what is unavailable in our setting, a global state estimate or external tracking~\cite{ahn2025relopush,Wu_2021,moura2022icra,dadoptis2025icra,lloyd2022tro}, while end-to-end visual RL~\cite{hafner2019icml,Sadeghi2017RSS,tobin2017domainrandomizationtransferringdeep,mnih2015nature,yarats2020improvingsampleefficiencymodelfree} needs neither but is sample-inefficient. Privileged teacher-student distillation~\cite{yin2025visualmimic,wu2025icra,uppal2024spin,lyu2025dywa,cheng2024icra,agarwal2022corl,chen2023sr,wang2024penspin} is the usual remedy, yet an omniscient teacher acts on information the student never observes, an instance of the \emph{imitation gap}~\cite{weihs2021advisor}: high teacher success alone does not guarantee a distillable teacher.

Rather than changing the imitation objective, EgoPush uses the teacher's observation function as an inductive bias: we train a privileged teacher under egocentric observability constraints. The teacher observes sparse object keypoints masked by a virtual field of view, and target-reference cues are revealed only when the anchor object lies in a central visible region. This design encourages circling and facing the anchor while pushing, behaviors that an egocentric visual student can recover; the gate is defined for anchor-relative goals (Sec.~\ref{subsec:phase1}).

Our contributions are: (i) a mobile multi-object rearrangement system that needs no global pose, map, or external tracking at deployment and transfers zero-shot to hardware, a regime no system in Table~\ref{tab:scope} operates in; (ii) the observation-constrained teacher that makes it learnable, and the finding behind it, that the observation function alone separates a $0\%$ student from an $87.3\%$ one; and (iii) two designs that make such a teacher trainable, a role-grouped object-centric interface shared by teacher and student, and stage-wise rewards with temporally decayed completion terms.

\section{Related Work}

\begin{table}[t]
    \centering
    \caption{Deployment Assumptions of Related Systems\tabnote{\emph{Onboard}: the policy consumes only robot-mounted sensing. \emph{No global}: no global pose, map, or external tracking at deployment. \emph{Multi}: more than one object rearranged. \emph{Horizon}: several sequential placements. 
    }}
    \label{tab:scope}
    \footnotesize
    \setlength{\tabcolsep}{3pt}
    \renewcommand{\arraystretch}{1.05}
    \begin{tabular}{l ccccc}
        \toprule
        \textbf{Method} & \textbf{Onboard} & \textbf{No global} & \textbf{Multi} & \textbf{Horizon} & \textbf{Real} \\
        \midrule
        ReloPush~\cite{ahn2025relopush}    & \xmark & \xmark & \cmark & \cmark & \cmark \\
        SAMaps~\cite{wu2020spatial}\,/\,SIM~\cite{Wu_2021} & \xmark & \xmark & \cmark & \cmark & \cmark \\
        DyWA~\cite{lyu2025dywa}            & \xmark & \cmark & \xmark & \xmark & \cmark \\
        SPIN~\cite{uppal2024spin}          & \cmark & \cmark & \xmark & \xmark & \cmark \\
        \rowcolor{oursblue}
        \textbf{EgoPush}                   & \cmark & \cmark & \cmark & \cmark & \cmark \\
        \bottomrule
    \end{tabular}
\vspace{-5mm}
\end{table}

\noindent\textbf{Non-prehensile Mobile Manipulation.}
Pushing lets mobile robots move objects that cannot be grasped, e.g., to navigate clutter, rearrange objects, or reconfigure poses~\cite{yang2025iros,dai2024interactive,ahn2025relopush,lyu2025dywa,jinag2023iros,dengler2022iros,bui2025push}.
Planning-based solutions combine contact models with search or force feedback~\cite{song2020iros,ren2025tro,liu2023ral,heins2024ral,ozdamar2024ral} and commonly assume reliable global state: known maps~\cite{wu2020spatial}, bird's-eye-view guidance~\cite{liu2025ais}, multi-agent coordination~\cite{tang2024rss,tang2026tro}, or all objects staying within the frustum~\cite{haramati2024entitycentric}.
Such assumptions break under narrow onboard views, frequent occlusions, and interaction-induced object motion. Table~\ref{tab:scope} sorts the closest systems by what they require at deployment: to our knowledge, every system that completes mobile rearrangement assumes a global pose estimate or a map, while the end-to-end policies that need neither (Table~\ref{tab:baseline_comparisons}) do not complete the task.

\noindent\textbf{Active Perception.}
Active perception, controlling sensing to acquire task-relevant information~\cite{bajcsy1988active,bajcsy2018revisiting}, is usually realized with a \emph{dedicated} perceptual degree of freedom such as a pan--tilt neck or a moving camera~\cite{cheng2018active,kerr2025eyerobot,xiong2025via,li2025actloc}.
A body-fixed camera on a pushing robot removes this separation: the same base motion must manipulate objects and redirect sensing, and occluded cues must be re-acquired repeatedly while maintaining contact.
Unlike single-object pushing or collision-reduction settings~\cite{hwang2026cura}, EgoPush studies this coupled setting \emph{without} a separate gaze controller or privileged global state (Sec.~\ref{sec:active_perception}). We use the visibility rule as an explicit inductive bias and measure whether the resulting look-to-act behavior is inherited by the student.

\begin{figure}[t]
    \centering
    \includegraphics[width=0.58\columnwidth]{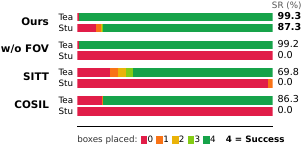}\hfill
    \includegraphics[width=0.40\columnwidth]{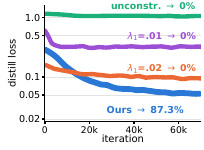}
    \caption{\textbf{Imitation-gap baselines on the full four-object cross.} Left: boxes placed ($4$ = success) and SR for each teacher (Tea) and its egocentric student (Stu). Right: student distillation loss under one pipeline (SITT with $\lambda_1\in\{0.01,0.02\}$).}
    \label{fig:imitation_gap}
\vspace{-5mm}
\end{figure}

\noindent\textbf{Teacher-Student Distillation.}
Learning control directly from pixels avoids explicit state estimation but is sample-inefficient and brittle under partial observability~\cite{hafner2019icml,mnih2015nature,Geles2024RSS}, motivating remedies such as differentiable simulation~\cite{you2025accelerating,heeg2025icra}.
A popular remedy is privileged-teacher-to-visual-student distillation~\cite{yin2025visualmimic,wu2025icra}: a teacher is first trained on privileged signals such as ground-truth states or noiseless geometry~\cite{uppal2024spin,agarwal2022corl,chen2023sr}, then transferred to a visual student by imitation~\cite{ross2011dagger}.
This assumes the teacher's behavior is \emph{recoverable} from the student's partial observations; in egocentric mobile manipulation an omniscient teacher can instead exploit off-screen information and leave insufficient visual evidence for a camera-based student, yielding ambiguous supervision.
This \emph{imitation gap}~\cite{weihs2021advisor} is typically addressed by adapting the training objective, either on the student side, e.g., an adaptive imitation reward toward the privileged expert (COSIL~\cite{nguyen2022cosil}), or on the teacher side, e.g., a penalty toward a jointly trained student (SITT~\cite{messikommer2025sitt}). EgoPush instead constrains the teacher's \emph{observation function} a priori, which needs no student-in-the-loop signal. On our task, this hard visibility design performs better than the two evaluated soft remedies (Fig.~\ref{fig:imitation_gap}).

\begin{figure*}[!t]
    \centering
    \includegraphics[width=0.94\textwidth]{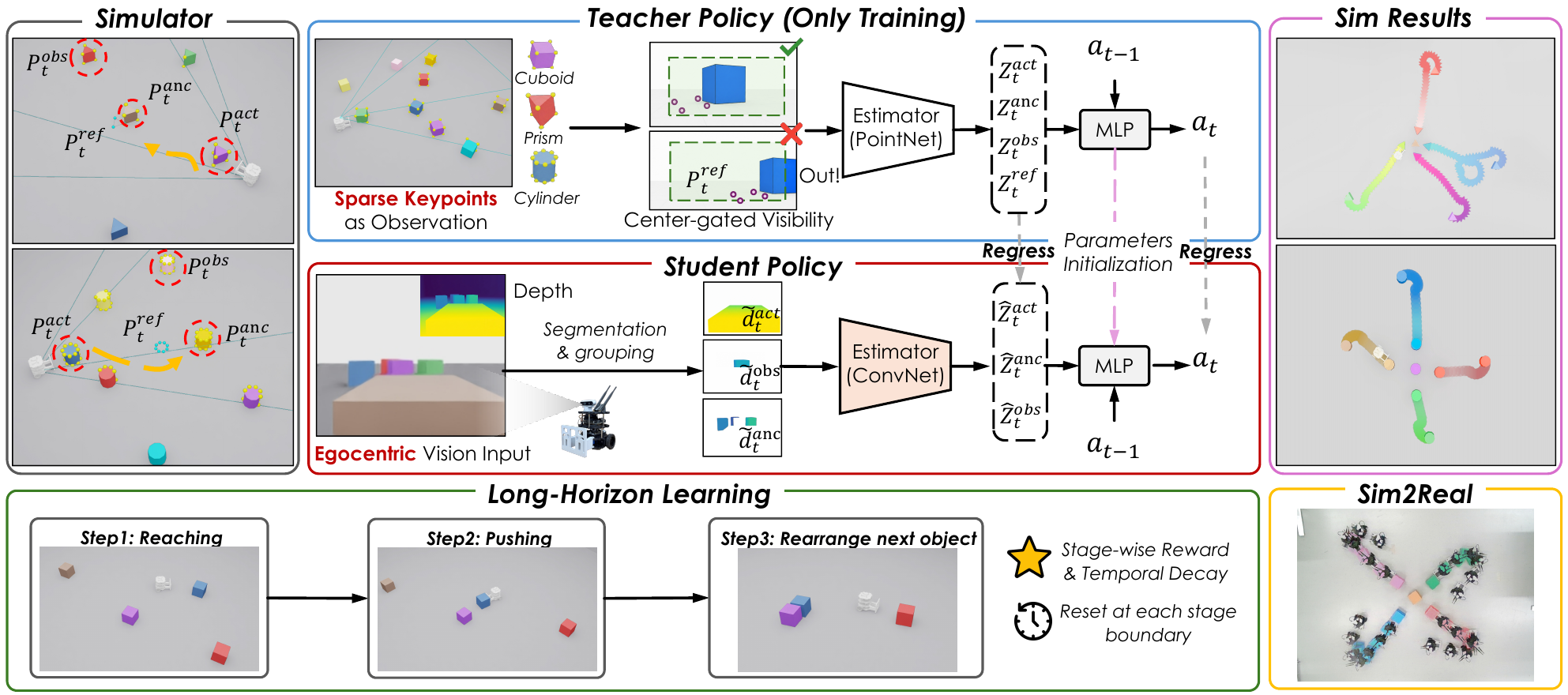}
    \caption{\textbf{EgoPush Overview}.
    A privileged teacher trains from visibility-limited sparse keypoints; the egocentric student consumes role-grouped depth (RGB used only for instance masks) and is distilled online via action plus relational losses, enabling zero-shot sim-to-real transfer.}
    \label{fig:framework}
\vspace{-5mm}
\end{figure*}

\section{Method}
EgoPush (Fig.~\ref{fig:framework}) builds on a role-based object-centric representation shared by teacher and student (Sec.~\ref{subsec:semantic}): Phase~1 trains the teacher by RL under constrained privileged observations (Sec.~\ref{subsec:phase1}), and Phase~2 distills it into an egocentric visual student by online imitation (Sec.~\ref{subsec:phase2}).

\subsection{Assumptions}
\label{subsec:assumptions}
The setting is deliberately bounded, and we state its givens once.
\emph{Perception}: objects are re-detected every frame by an external segmentation front-end; no tracker, global pose, map, or off-screen information is used.
\emph{Roles}: the task binds each object identity (here its color) to a position in the stage order, so role masks are routed by identity, never per trial or by a person.
\emph{Goals}: the order of active objects and each target slot (a fixed anchor-relative offset) are task-defined. Within these givens the deployed policy sees only egocentric depth, receives no goal input, and infers the current slot from the anchor, the placed objects, and a stage counter it maintains itself (Sec.~\ref{subsec:phase2}).

\subsection{Object-Centric Latent Representation}
\label{subsec:semantic}
At each stage, visible task instances are assigned to one of three task roles: the currently manipulated object (\textbf{active}), the object whose pose defines the local target (\textbf{anchor}), and every remaining scene object (\textbf{obstacle}).
We encode the three groups with a shared-weight set encoder and concatenate the resulting per-group latents into a single \textbf{object-centric latent representation}.
Shared weights place all groups in a common embedding space, so the policy reasons over \emph{relative spatial relations between group latents}; teacher and student share this partition.

\subsection{Phase~1 (Teacher): Online Reinforcement Learning}
\label{subsec:phase1}
\noindent\textbf{Action.}
The differential-drive policy outputs $a_t=[v_t, \omega_t]$ (linear, angular velocity), converted to wheel velocities by a PD controller.

\noindent\textbf{Privileged keypoints.}
The teacher observes sparse keypoints~\cite{wu2026precise,wu2025icra,uppal2024spin} for active/anchor/obstacle plus a target reference set $P_t^{\text{ref}}$ sampled from the predefined target configuration; for each task category the set of anchor-relative slots is fixed and stage $s$ uses a fixed offset, so $P_t^{\text{ref}}$ acts as a consistent pose-alignment cue.

\noindent\textbf{Constrained observations.}
To keep teacher trajectories visually recoverable for the student, we apply: (1) \textit{virtual egocentric FOV masking} that masks points outside a robot-pose-based viewing frustum or beyond a maximum range; (2) \textit{center-gated visibility for $P_t^{\text{ref}}$}: visible only when the anchor lies within the FOV \emph{and} inside a central gated region $(\pm u_{\text{gate}}, \pm v_{\text{gate}})$ in normalized image coordinates -- otherwise $P_t^{\text{ref}}$ is masked.
The teacher then observes $\{P_t^{\text{act}}, P_t^{\text{anc}}, P_t^{\text{obs}}, P_t^{\text{ref}}\}$ plus $a_{t-1}$, with $\theta_h{\times}\theta_v{=}70^\circ{\times}60^\circ$, $d_{\max}{=}5$\,m, $u_{\text{gate}}{=}0.85$ and $v_{\text{gate}}{=}1.0$, so the gate is effectively horizontal, and obstacles kept as a distance-sorted padded set of up to $32$ groups. The mask covers out-of-view and out-of-range cues but not inter-object occlusion (Sec.~\ref{sec:limitation}).

\noindent\textbf{Rewards.}
(1) \textit{Stage-aligned completion reward.} 
Each object-wise stage is decomposed into a \emph{reach}$\rightarrow$\emph{place}
sequence. Since sparse terminal success alone cannot distinguish efficient and
inefficient completions, we introduce a stage-aligned completion weight
$\eta_t=\mathrm{clip}\big((T_s-\tau_t)/(T_s+\epsilon_0),0,1\big)$, where $\tau_t$ is the elapsed step count
within the current stage and resets at every stage boundary, keeping completions non-negative past $T_s$. The reach and place
completion rewards, weighted $10$ and $200$, are
\begin{subequations}
\label{eq:completion_rewards}
\begin{align}
r_{\mathrm{reach},t}
&= \eta_t\,
\mathbb{I}\!\left[
\|\mathbf{p}_{\mathrm{rbt},t}-\mathbf{p}_{\mathrm{act},t}\|_2
< \varepsilon_{\mathrm{reach}}
\right],
\label{eq:r_reach}\\[-0.5mm]
r_{\mathrm{place},t}
&= \eta_t\,
\mathbb{I}\!\left[
\begin{array}{l}
\|\mathbf{p}_{\mathrm{act},t}-\mathbf{p}_{\mathrm{ref},t}\|_2
< \varepsilon_{\mathrm{align}}, \text{and}\\[-0.5mm]
|\Delta\psi_t| < \varepsilon_{\Phi},\text{and} \quad
\|\mathbf{v}_{\mathrm{act},t}\|_2 < \varepsilon_v
\end{array}
\right].
\label{eq:r_place}
\end{align}
\end{subequations}
\noindent where $\mathbf{p}_{\mathrm{rbt},t}$, $\mathbf{p}_{\mathrm{act},t}$, and
$\mathbf{p}_{\mathrm{ref},t}$ denote the robot, active-object, and reference
target positions; $\Delta\psi_t$ is the active-object yaw error; and
$\mathbf{v}_{\mathrm{act},t}$ is the active-object planar velocity, with $\varepsilon_{\mathrm{reach}}{=}0.25\,$m.
(2) \textit{Progress shaping (Phase-Gated Distance Decrease)}:
\begin{equation}
\begin{aligned}
r_{\text{dist},t} =\;& w_{\text{rbt}}(d_{\text{rbt},t-1}-d_{\text{rbt},t})\,\mathbb{I}[g_t{=}0] \\
&+ w_{\text{ref}}(d_{\text{ref},t-1}-d_{\text{ref},t})\,\mathbb{I}[g_t{=}1],
\end{aligned}
\label{eq:r_dist}
\end{equation}
\noindent where $g_t\in\{0,1\}$ is a phase gate that switches from \emph{reach} ($g_t{=}0$) to \emph{place} ($g_t{=}1$) at the reach event, $d_{\text{rbt},t}=\|\mathbf{p}_{\text{rbt},t}-\mathbf{p}_{\text{act},t}\|_2$, and $d_{\text{ref},t}=\|\mathbf{p}_{\text{act},t}-\mathbf{p}_{\text{ref},t}\|_2$, with $w_{\text{rbt}}{=}1$ and $w_{\text{ref}}{=}2.5$.
(3) \textit{Smoothness}: penalizes consecutive-action changes, weighted $-2{\times}10^{-4}$.
(4) \textit{Slowdown Near Target}: rewards low object speed near the reference, weighted $0.05$. The four terms are summed under a global scale of $10$.

\noindent\textbf{Termination.}
Episodes end on task success, world-box exit, robot-obstacle / active-obstacle collision, or the episode step limit; early termination replaces a dense collision penalty to avoid reward hacking.

\noindent\textbf{EgoPush-Teacher.}
We use PointNet~\cite{qi2017pointnet} to encode each visible semantic point set
$P_t^k$ into a permutation-invariant latent representation, where
$k\in\mathcal{K}=\{\mathrm{act},\mathrm{anc},\mathrm{obs},\mathrm{ref}\}$.
The teacher policy predicts actions from the group-wise latents and the previous action:
\begin{equation}
\begin{aligned}
Z_t^k &= \mathrm{PointNet}_{\theta}(P_t^k), \quad k\in\mathcal{K}, \\
a_t &= \mathrm{MLP}\!\left(a_{t-1},\, Z_t^{\mathrm{act}},\, Z_t^{\mathrm{anc}},\, Z_t^{\mathrm{obs}},\, Z_t^{\mathrm{ref}}\right).
\end{aligned}
\label{eq:teacher_policy}
\end{equation}
Group encoders are shared-weight PointNets ($64$, $128$, then $128{\to}64{\to}16$) feeding a two-layer $256$-wide trunk. We train with PPO~\cite{schulman2017ppo} on $8{,}192$ environments (rollout $32$, $8$ epochs, clip $0.2$, KL-adaptive lr from $2.5{\times}10^{-4}$), randomizing box mass $[0.4,0.6]\,$kg and friction $[0.3,0.7]$.

\subsection{Phase~2 (Student): Supervised Learning}
\label{subsec:phase2}
\noindent\textbf{Observation and task interface at deployment}.
The student carries an egocentric RGB-D camera; RGB is used \emph{only} to obtain per-instance masks, and the policy consumes depth segments. Under these assumptions, the roles resolve as follows: at stage $s$ the $s$-th object is \emph{active}, the \emph{anchor} is one fixed object (cross) or the previously placed object (line, Sec.~\ref{sec:validation}), and all other objects, placed or not, are \emph{obstacles}. The teacher receives the stage's target slot as $P_t^{\text{ref}}$, whereas the student receives no goal input and must infer it from the anchor, the already-placed objects (observed as obstacles), and its own stage counter.
\emph{Stage tracking}: the student advances that counter from a completion signal it emits itself. Throughout distillation the role routing and the teacher's target slot follow the student's \emph{own} counter, never the hidden stage, so the DAgger loop labels the student's own counter errors. Ground-truth completion only schedules six zero-action steps, executed with the role masks held pre-switch, which teach the student to hold still exactly when the observed active object sits at its slot; at deployment, four consecutive zero-action frames ($|v_t|{<}0.02\,$m/s and $|\omega_t|{<}0.05\,$rad/s) advance the counter and re-route the role masks, after which the new active object is off its slot and motion resumes. No privileged stage signal is read at evaluation or on hardware; Sec.~\ref{sec:validation} measures how well this self-maintained counter tracks the hidden stage index.

Concretely, instance segmentation $S_{\text{inst}}(\cdot)$ yields masks $\{M_t^{(i)}\}_{i=1}^{N_t}$, each assigned to a role group $\mathcal{I}_t^{k}$, and masked depth is aggregated per group:
\begin{equation*}
\tilde{d}_t^{k} = \sum_{i \in \mathcal{I}_t^{k}} M_t^{(i)} \odot d_t,\quad k \in \{\mathrm{act}, \mathrm{anc}, \mathrm{obs}\}.
\end{equation*}
\begin{figure}[t]
  \centering
  \includegraphics[width=0.95\columnwidth]{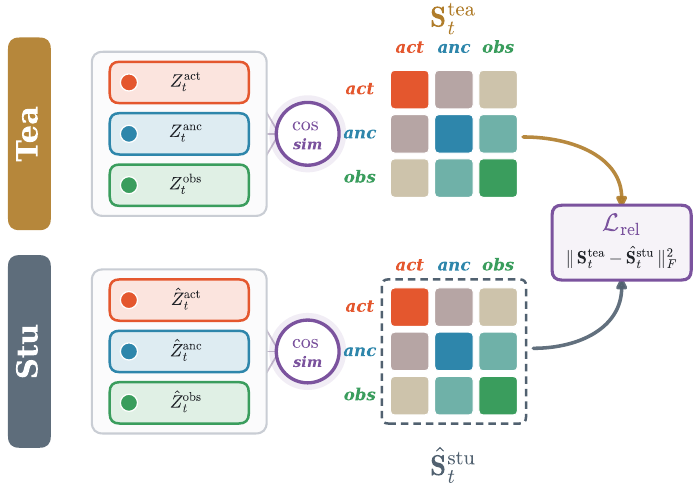}
  \caption{\textbf{Relational Distillation.} The pairwise cosine-similarity matrix over the shared act/anc/obs groups is matched between teacher and student (Eq.~\ref{eq:distill_loss}).}
  \label{fig:pairwise_relation}
\vspace{-5mm}
\end{figure}

Disjoint masks make these three role-grouped depth layers interference-free.
On hardware we instantiate $S_{\text{inst}}(\cdot)$ with two front-ends sharing this role-grouping interface: HSV thresholding, and color-gated proposals refined by box-prompted \textit{MobileSAM}~\cite{mobile_sam} (Sec.~\ref{sec:real_world}); zero-shot segmenters~\cite{ravi2024sam2segmentimages,simeoni2025dinov3} could replace the color gating in general scenes.
For the depth sim-to-real gap we simulate the sensor's noise pattern in training and inpaint real depth~\cite{bertalmio2001ns,zhang2026highspeedvisionbasedflightclutter} at deployment. This choice balances latency against fidelity among four treatments of the raw D435i depth that we compared on hardware: a learned RGB-guided denoiser~\cite{liu2025manipulation} preserves the most surface detail but costs about $50$\,ms per frame even on the server; the onboard hole-filling, spatial, and temporal filter stack of~\cite{zhuang2023corl} takes about $15$\,ms on the Jetson and still flickers at close range and loses top surfaces at distance; filling each mask with its median depth is stable and cheap but discards all intra-object geometry. Navier--Stokes inpainting within each mask keeps the median fill's stability and cost (about $2$\,ms per frame) while retaining more geometry, at a comparable success rate.

\noindent\textbf{EgoPush-Student.}
The student replaces the teacher's PointNet encoders with a CNN over each masked depth layer and keeps the teacher's MLP trunk and action head: the weights are copied and the first-layer columns acting on $Z_t^{\text{ref}}$ are dropped.
Let $\mathcal{K}_{\mathrm{stu}}=\{\mathrm{act},\mathrm{anc},\mathrm{obs}\}$ denote the visible semantic groups used by the student:
\begin{equation}
\begin{aligned}
\hat{Z}_t^k &= \mathrm{CNN}_{\phi}\!\left(\tilde{d}_t^k\right), \quad k\in\mathcal{K}_{\mathrm{stu}}, \\
\hat{a}_t &= \mathrm{MLP}\!\left(a_{t-1},\, \hat{Z}_t^{\mathrm{act}},\, \hat{Z}_t^{\mathrm{anc}},\, \hat{Z}_t^{\mathrm{obs}}\right).
\end{aligned}
\label{eq:student_policy}
\end{equation}
The stage counter is \emph{not} an input to Eq.~\ref{eq:student_policy}: it only re-routes the role masks, changing the \emph{contents} of the three depth layers rather than the input dimension. At stage $s$ the $s$-th object enters $\tilde{d}_t^{\mathrm{act}}$ and the placed ones join $\tilde{d}_t^{\mathrm{obs}}$ (the last one $\tilde{d}_t^{\mathrm{anc}}$ in the line formation), so the slot follows from the anchor-relative geometry.
Each layer passes a CNN ($32$, $64$, $64$ channels, then MLP $256{\to}16$). Offline BC is brittle here, since early deviations leave the teacher distribution, so we distill online, DAgger-style with $512$ environments and Adam at $3{\times}10^{-4}$, querying the teacher at the states the student visits, and additionally randomize the camera mount ($\le0.14\,$m, $\pm2^\circ$) and drop $10\%$ of depth pixels, edge-weighted, with median fill.

The teacher's latents were shaped jointly with the privileged $Z_t^{\text{ref}}$, so element-wise matching would transfer features whose meaning depends on an input the student never sees.
As a lighter auxiliary constraint, we align only the \emph{relational structure} among the shared groups $\mathcal{K}_{\text{shared}}{=}\{\mathrm{act},\mathrm{anc},\mathrm{obs}\}$ through their pairwise cosine-similarity matrix $\mathbf{S}$ (Fig.~\ref{fig:pairwise_relation}):
\begin{equation}
\begin{aligned}
S_{i,j,t} &= \frac{Z_t^i \cdot Z_t^j}{\|Z_t^i\|_2 \, \|Z_t^j\|_2}, \quad i,j \in \mathcal{K}_{\text{shared}}, \\
\mathcal{L}_t &= \lambda_a \, \| a_t^{T} - \hat{a}_t \|_2^2 + \lambda_{\text{rel}} \, \| \mathbf{S}_t^{\text{tea}} - \hat{\mathbf{S}}_t^{\text{stu}} \|_F^2
\end{aligned}
\label{eq:distill_loss}
\end{equation}

\noindent where $a_t^{T}$ is the DAgger teacher label and $(\lambda_a, \lambda_{\text{rel}})$ are loss weights ($\lambda_a{=}1$, $\lambda_{\text{rel}}{=}0.2$).
Since $\mathbf{S}$ is symmetric with unit diagonal, the term constrains three pairwise relations among the shared latents, and it is part of the distillation objective in every experiment. Sec.~\ref{sec:ablation} ablates it on task success.
Starting from the copied teacher weights, we apply additional camera-pose randomization on top of Phase~1 physical randomization to improve sim-to-real robustness.

\section{Experiments}
\label{sec:exp_setup}
We evaluate EgoPush in simulation (NVIDIA Isaac Lab) and in the real world, asking: (i) why end-to-end RL and imitation baselines fail; (ii) whether restricting the teacher's observations yields more distillable supervision; (iii) whether stage-aligned, temporally decayed rewards speed up long-horizon RL; and (iv) whether the student transfers zero-shot to a real TurtleBot.

We use cross-shape rearrangement (Fig.~\ref{fig:head}) as the primary evaluation task throughout, baselines included, and add a \emph{simplified two-object setup}, stage one of that task, to separate methods that all read near zero on the full cross.
Each policy is trained per task category on cuboid active objects. A Unity port of the cross task is on the project page.
We report \textbf{SR} (placement success), \textbf{Reach} (first contact with active object), \textbf{Time} (policy steps per successful episode), and \textbf{Traj.} (robot path length).
We train in Isaac Sim~4.5 and Isaac Lab~2.1 on one NVIDIA A6000; success requires the active object within $\varepsilon_{\text{align}}{=}0.10\,$m and $\varepsilon_{\Phi}{=}0.20\,$rad of its reference and at rest ($\|\mathbf{v}_{\text{act}}\|_2{<}0.05\,$m/s), with a per-stage decay horizon $T_s{=}1000$ steps, which is also the episode limit ($100\,$s at $10\,$Hz). Each policy uses one training seed and $1{,}024$ evaluation episodes unless a table states otherwise (ACT in Fig.~\ref{fig:act_diagnosis} uses four seeds and $512$ episodes per point), so the reported intervals bound episode-sampling noise and not between-seed variation.

\subsection{Baseline Comparisons}
\begin{table}[t]
    \centering
    \caption{Baseline Comparison\tabnote{\emph{Stages}: mean completed placements (of $4$). N/A: fewer than $1\%$ successes, too few for a trajectory statistic. E2E-RGB and E2E-RGBD are omitted for space (Reach $5.57$/$6.54$, SR $0.00$/$0.29$). Wilson $95\%$ half-widths at $n{=}1024$: ${\pm}0.4$ near $0$, ${\pm}2.4$ at $18\%$, ${\pm}2.0$ at $87\%$; $0/1024$ bounds SR below $0.37\%$.}}
    \label{tab:baseline_comparisons}
    \footnotesize
    \setlength{\tabcolsep}{2pt}
    \renewcommand{\arraystretch}{0.95}
    \begin{tabularx}{\columnwidth}{L L C C C}
        \toprule
        \multicolumn{5}{c}{\textbf{Two-object setup (stage one of the cross task)}} \\
        \midrule
        \textbf{Method} & \textbf{Input} & \textbf{Reach$\uparrow$} & \textbf{SR$\uparrow$} & \textbf{Traj.$\downarrow$} \\
        \midrule

        SIM~\cite{Wu_2021} & RGB-D+map+odom & 30.94 & 19.26 & 9.65 \\
        \midrule

        \rowcolor{sectiongray}
        \multicolumn{5}{l}{\textit{End-to-end RL}}\\
        RGBD-Sem & RGB-D+seg  & 39.12 & 0.10 & N/A   \\
        E2E-Curr.& RGB-D      & 32.75 & 0.78 & N/A   \\
        RGBD-RNN & RGB-D      & 15.43 & 0.10 & N/A   \\
        \midrule

        \rowcolor{sectiongray}
        \multicolumn{5}{l}{\textit{Behavior cloning (BC)}}\\
        ACT (Tele, 50) & RGB-D & 1.76  & 0.00  & N/A    \\
        ACT (RL, 50)  & RGB-D & 55.47 & 8.69  & 11.29 \\
        ACT (RL, 500) & RGB-D & 63.09 & 18.16 & 11.91 \\
        \midrule

        \rowcolor{oursblue}
        \textbf{Ours} & \textbf{RGB-D+seg} & \textbf{100.00} & \textbf{100.00} & \textbf{4.66} \\
        \midrule[\heavyrulewidth]
        \multicolumn{5}{c}{\textbf{Full four-object cross}} \\
        \midrule
        \textbf{Method} & \textbf{Input} & \textbf{Reach$\uparrow$} & \textbf{SR$\uparrow$} & \textbf{Stages$\uparrow$} \\
        \midrule
        SIM~\cite{Wu_2021} & RGB-D+map+odom & 10.4 & 0.00 & 0.00 \\
        ACT (RL, 1k) & RGB-D & 72.1 & 0.00 & 0.10 \\
        \rowcolor{oursblue}
        \textbf{Ours} & \textbf{RGB-D+seg} & \textbf{95.6} & \textbf{87.3} & \textbf{3.57} \\
        \bottomrule
    \end{tabularx}
\end{table}

\begin{figure}[!t]
    \centering
    \includegraphics[width=\columnwidth]{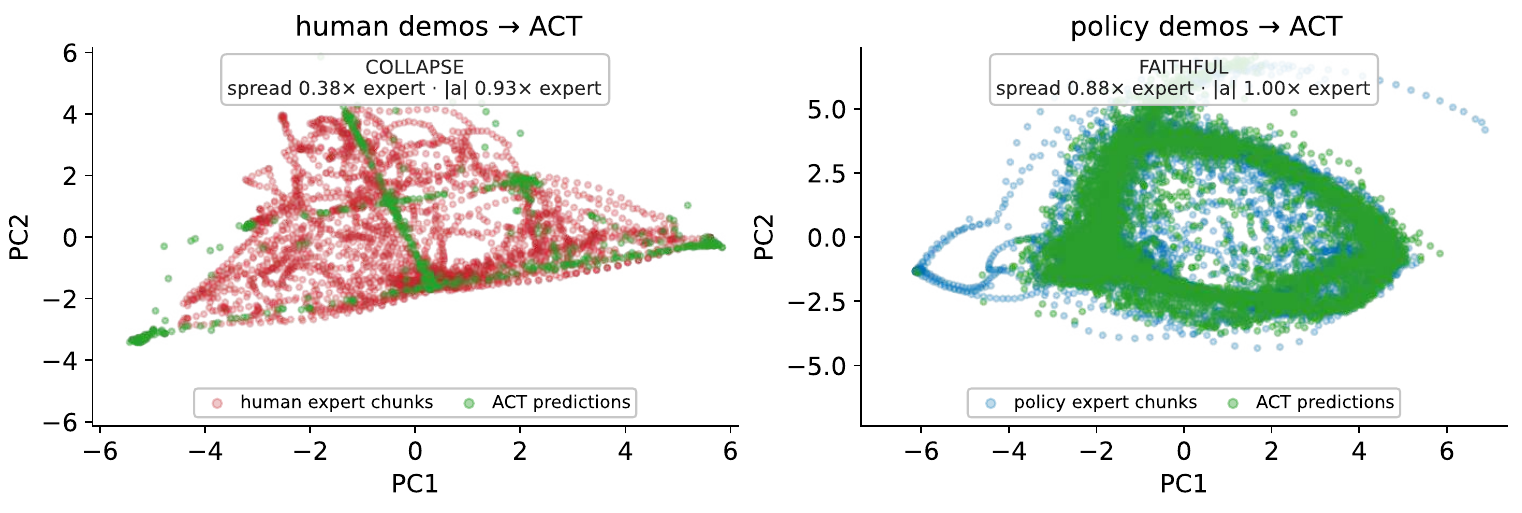}
    \caption{\textbf{Behavior cloning struggles in EgoPush.} Action-chunk PCA collapses on human teleop demos \textbf{(left)} and is faithful but insufficient on RL demos \textbf{(right)}.}
    \label{fig:act_why}
\vspace{-5mm}
\end{figure}

\label{sec:baseline_comparison}

\noindent\textbf{Two-Object Benchmark.}
We evaluate EgoPush against Spatial Intention Maps (SIM)~\cite{Wu_2021}, five end-to-end (E2E) visual RL variants (input modality, oracle masks, reach-first curriculum, or a GRU head), and behavior cloning (ACT~\cite{zhao2023learning} on human teleoperation vs.\ synthetic RL demonstrations) on the simplified two-object setup under matching egocentric sensing.
All methods share the environment and stage schedule; RL baselines also share our simulator timing and action bounds, the E2E variants use our reward set, and RGBD-Sem follows our role-grouped masks. The E2E variants are not held to the teacher's budget: each is trained until its return saturates, so their sub-$1\%$ SR is not a matter of training length. ACT keeps its original design, raw RGB-D with the previous action, and trains offline on a fixed demonstration set while our student queries the teacher online. SIM keeps its robot-centric map, waypoint action, and reward, but this setting supplies no ground-truth pose, so the map must be registered against an onboard estimate; we use integrated executed velocities, and the pushing itself drives the drift, since every contact moves the scene the map is registered against.
All E2E baselines stay below $1\%$ Success Rate (SR) despite reasonable Reach (Table~\ref{tab:baseline_comparisons}), so oracle masks alone (RGBD-Sem) do not solve long-horizon reasoning under contact.
SIM reaches $19.26\%$ SR, whereas EgoPush achieves $100\%$ Reach and SR.

BC via ACT~\cite{zhao2023learning} also struggles; to diagnose why, we compare teleoperation with synthetic RL demonstrations collected by our policy.
Teleoperation is costly ($71$ episodes collected, $50$ for training) and a poor signal for deterministic BC: with $50$ demos ACT obtains $0\%$ SR on teleop data versus $8.7\%$ on RL demos. 
Scaling RL demos improves SR monotonically but only to $18.16\%$ with $500$ demos (Fig.~\ref{fig:act_diagnosis}, left). 
Thus, RL demos partially mitigate teleoperation's failure, but the evaluated offline BC pipeline remains far below our online RL-to-vision distillation ($100\%$ SR).

\noindent\textbf{Full Cross-Task Benchmark.}
The bottom block of Table~\ref{tab:baseline_comparisons} extends the comparison to the full four-active-object cross, of which the two-object setting is stage one, so a method failing there cannot complete the sequence. Since end-to-end visual RL already stays below $1\%$ SR at stage one, we evaluate the remaining families under one shared protocol: map-based SIM and demonstration-driven ACT ($1{,}000$ RL demonstrations). Both reach $0\%$ SR and complete at most $0.1$ of the $4$ stages, whereas the EgoPush student completes $3.57$ stages with $87.3\%$ SR (teacher: $99.31\%$).

\noindent\textbf{Imitation-Gap Baselines.}
SITT~\cite{messikommer2025sitt} and COSIL~\cite{nguyen2022cosil} encourage teacher--student agreement through a teacher-side penalty and an adaptive student imitation reward, respectively, whereas EgoPush enforces visibility constraints on the teacher's observations. On the full four-object cross, SITT and COSIL students score $0\%$, as does the student of an unconstrained teacher ($99.22\%$, Sec.~\ref{sec:ablation}); only the constrained teacher ($99.31\%$) yields a usable student ($87.3\%$). Under these adaptations (COSIL ported to PPO, SITT swept over $\lambda_1$), constraining the observation function is the more effective intervention.

\begin{table}[t]
    \centering
    \caption{Teacher Observation Ablation\tabnote{Top: behaviors; bottom: teacher/student metrics, including the deployed student's \emph{active gain} $g$ (Sec.~\ref{sec:active_perception}). Student-SR Wilson $95\%$ intervals ($n{=}1024$) are disjoint: $[85.1,89.2]$, $[51.7,57.8]$, $[0,0.37]$.}}
    \label{tab:abl_distill_design_v2}
    \footnotesize
    \setlength{\tabcolsep}{0pt}
    \renewcommand{\arraystretch}{1.0}
    \begin{tabular*}{\columnwidth}{@{}l @{\hspace{3pt}}l @{\extracolsep{\fill}} c c c@{}}
        \toprule
        & \textbf{Behavior} & \textbf{Ours} & \textbf{w/o C-GV} & \textbf{w/o FOV} \\
        \midrule

        & &
        \raisebox{-0.5\height}{\includegraphics[width=0.27\columnwidth]{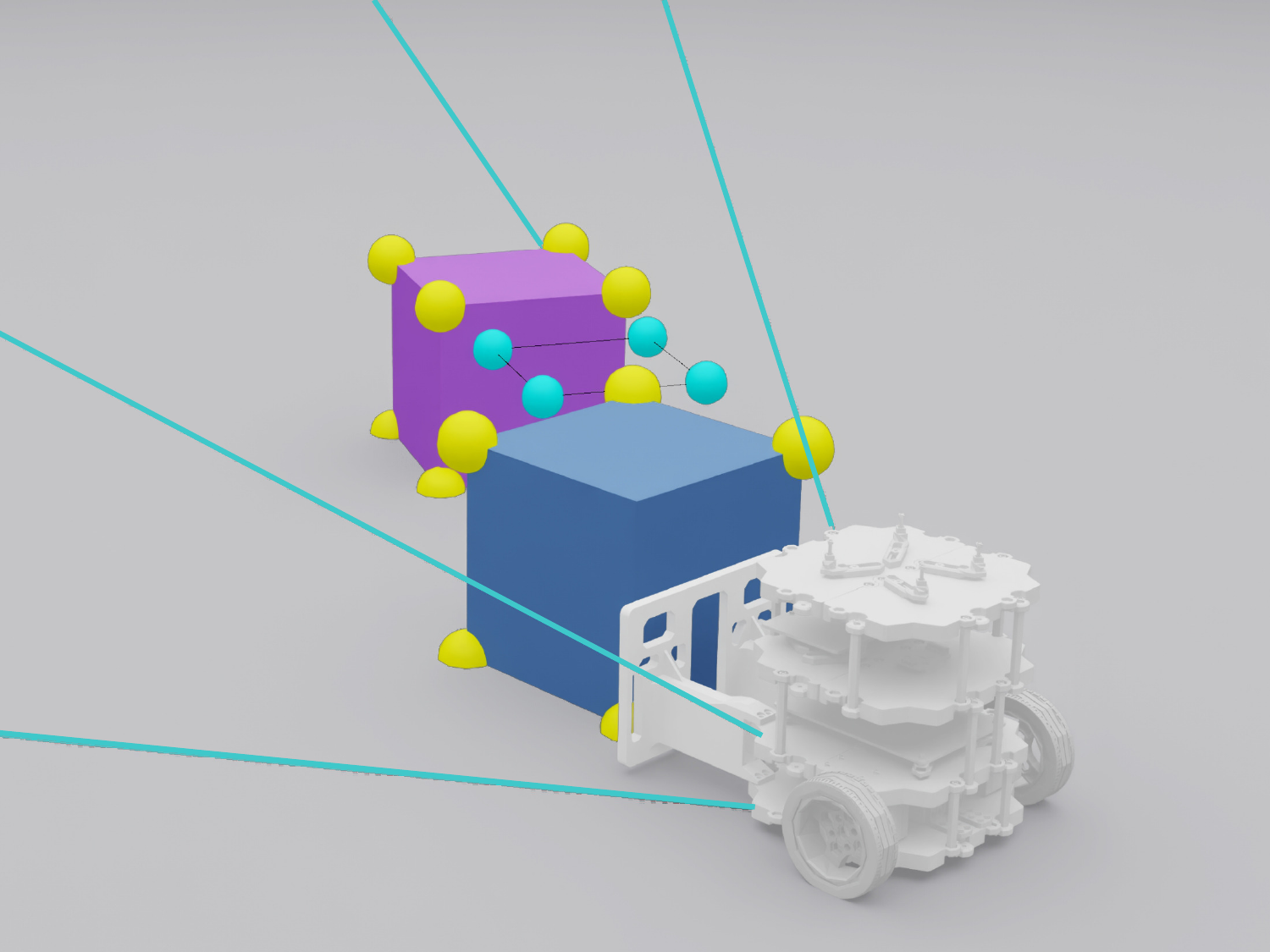}} &
        \raisebox{-0.5\height}{\includegraphics[width=0.27\columnwidth]{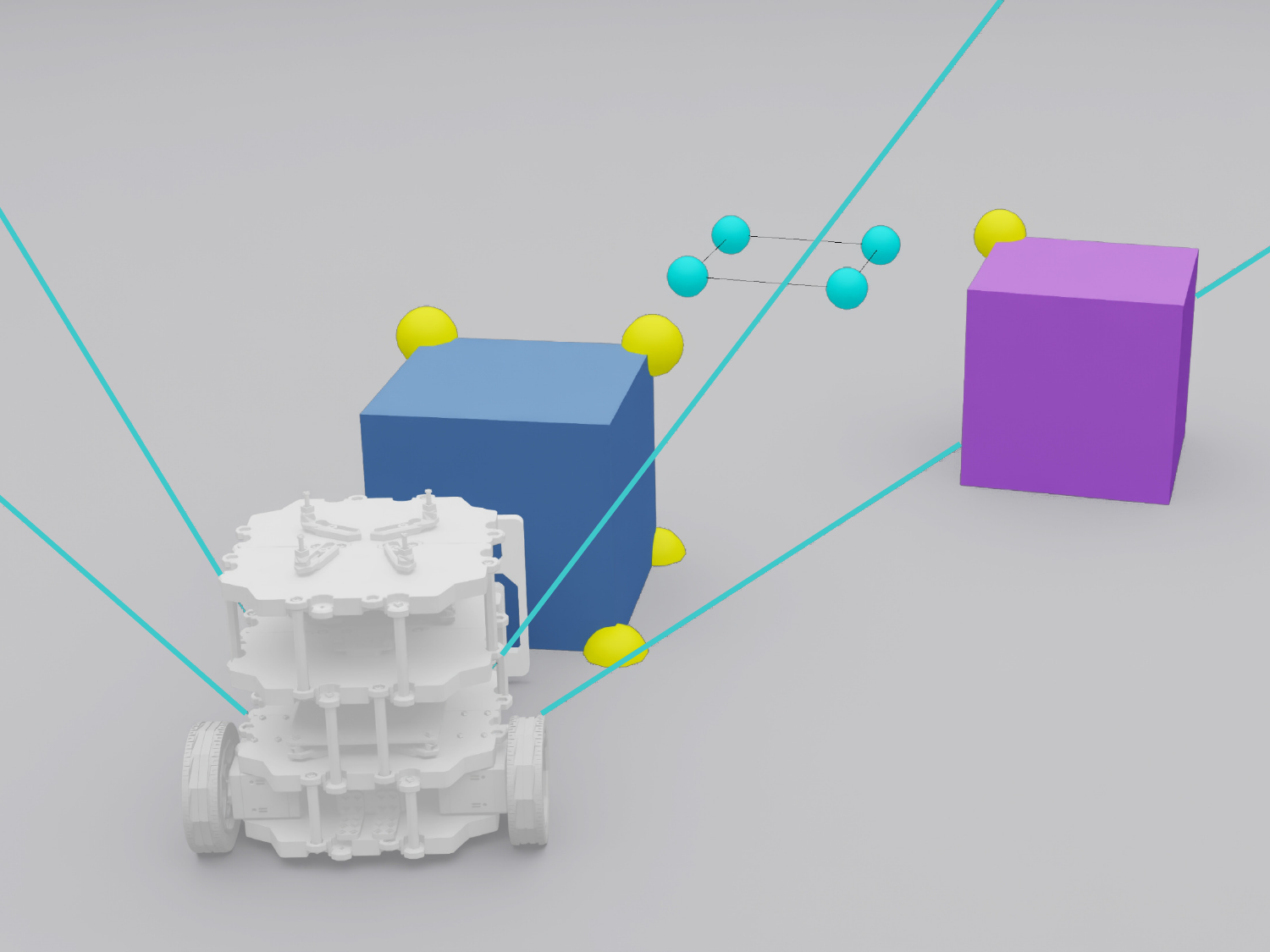}} &
        \raisebox{-0.5\height}{\includegraphics[width=0.27\columnwidth]{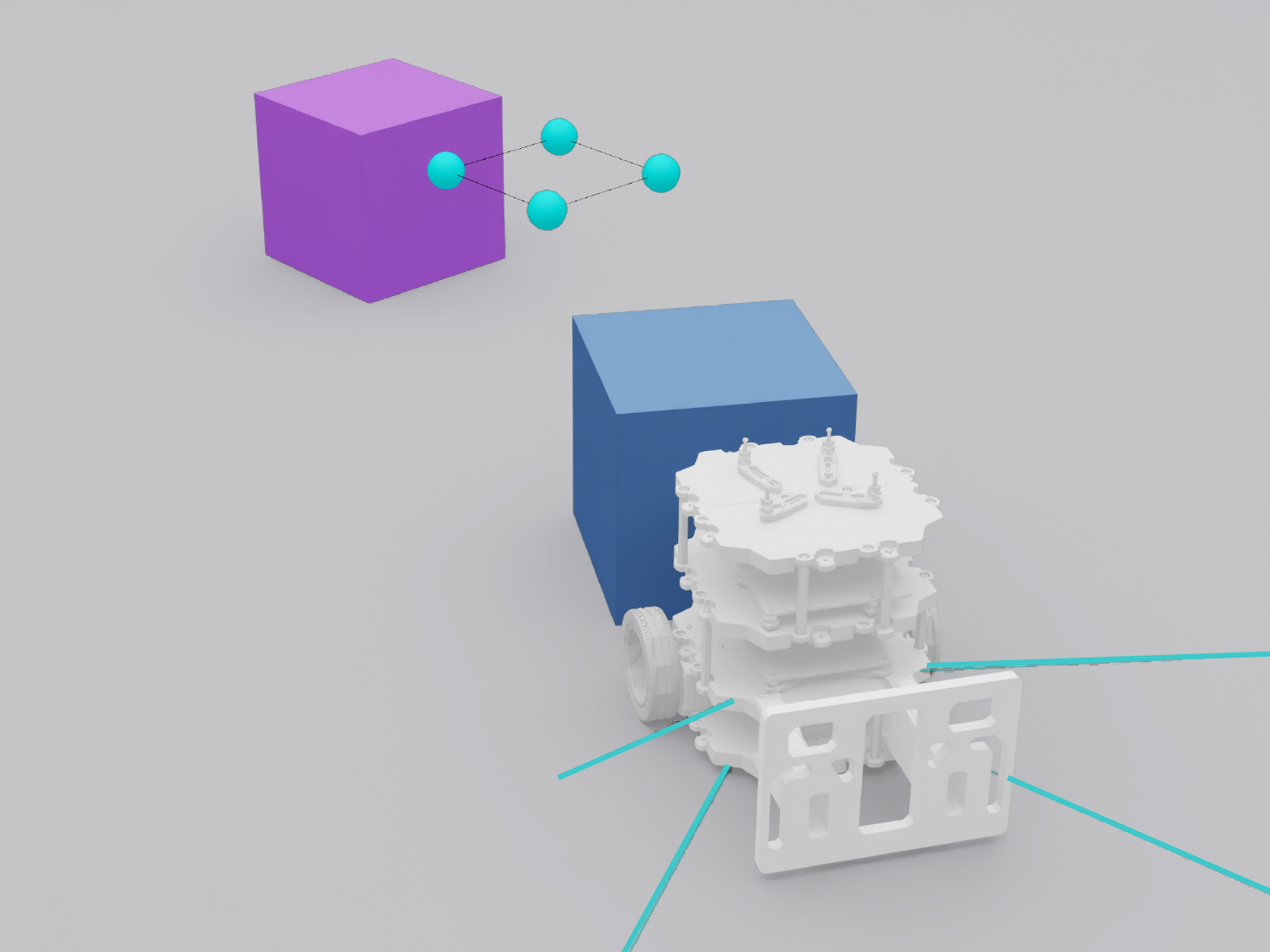}} \\
        \addlinespace[2pt] 
        
        \midrule
        \multirow{3}{*}{\rotatebox[origin=c]{90}{\scalebox{0.85}{\textbf{\color{gray!100}Teacher}}}} 
        & SR$\uparrow$   & \textbf{99.31} & 98.34 & 99.22 \\
        & Time$\downarrow$ & 395.59 & 394.39 & \textbf{373.20} \\
        & Traj.$\downarrow$  & 11.59 & 11.88 & \textbf{11.38} \\
        
        \midrule
        \multirow{4}{*}{\rotatebox[origin=c]{90}{\scalebox{0.85}{\textbf{\color{gray!100}Student}}}}
        & SR$\uparrow$   & \textbf{87.30} & 54.79 & 0.00 \\
        & Time$\downarrow$ & 431.99 & 419.65 & N/A \\
        & Traj.$\downarrow$  & \textbf{11.41} & 11.78 & N/A \\
        & $g\uparrow$   & \textbf{+0.59} & +0.36 & $-0.19$ \\
        \bottomrule
    \end{tabular*}
\end{table}

\begin{table}[t]
    \centering
    \caption{Credit-Assignment Ablation on Cross\tabnote{Final performance: \textbf{Ours} after ${\sim}45$k vectorized steps, the ablations after ${\sim}90$k. Time and Traj.\ are computed over successful episodes only.}}
    \label{tab:credit_assignment}
    \footnotesize
    \setlength{\tabcolsep}{8pt}
    \renewcommand{\arraystretch}{1.05}
    \begin{tabular}{l c c c}
        \toprule
        \textbf{Method} & \textbf{SR}$\uparrow$ & \textbf{Time}$\downarrow$ & \textbf{Traj.}$\downarrow$ \\
        \midrule
        Base    & 16.02 & 1076.42 & 13.78 \\
        SWR     & 87.50 &  505.13 & 14.24 \\
        SWR+TD  & 97.95 &  455.46 & 13.77 \\
        \rowcolor{gray!10}
        \textbf{Ours}   & \textbf{98.63} &  \textbf{443.26} & \textbf{12.83} \\
        \bottomrule
    \end{tabular}
\vspace{-3mm}
\end{table}

\begin{figure}[t]
    \centering
    \includegraphics[width=0.95\columnwidth]{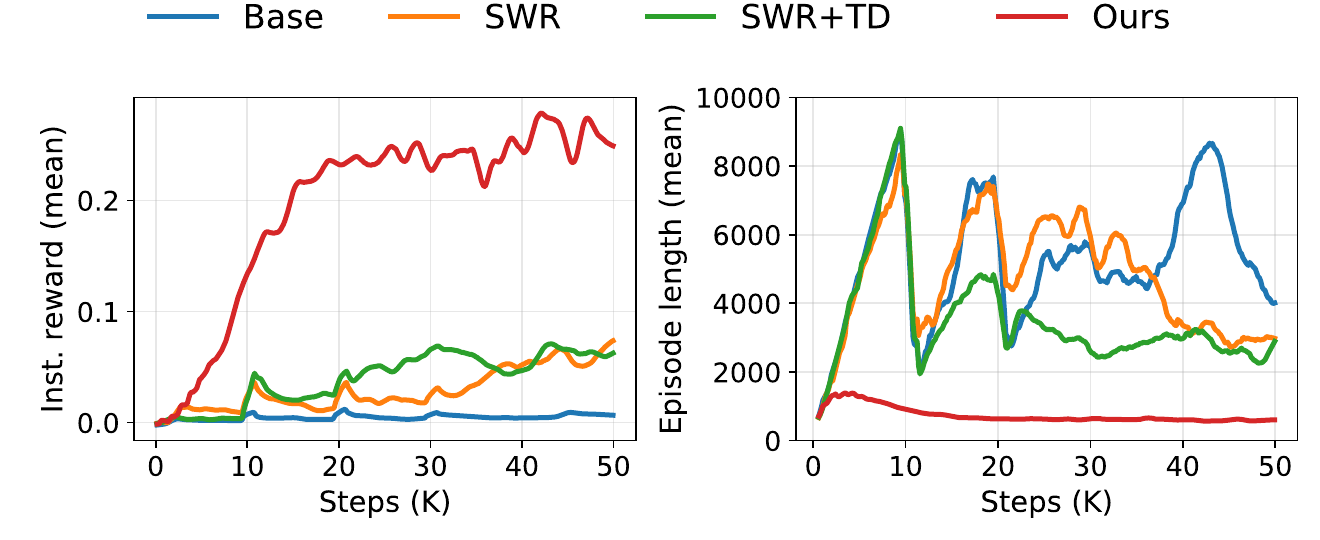}
    \caption{\textbf{Credit-assignment ablation on Cross.} First 50k vectorized steps ($8{,}192$ environments each). Left: per-variant instantaneous reward (scales differ); right: episode length.}
    \label{fig:credit_assignment}
\end{figure}

\subsection{Ablation Studies}
\label{sec:ablation}

\noindent\textbf{Restricted Observations for RL Teacher.}
The constrained teacher combines two designs: virtual egocentric FOV masking and center-gated visibility (C-GV) for privileged reference keypoints.
We compare against two relaxed variants: \textbf{w/o FOV} removes the FOV mask and exposes global keypoints, while \textbf{w/o C-GV} keeps the FOV but removes the center-gated visibility on the $P_t^{\text{ref}}$.
All variants share the algorithm, reward, hyperparameters, and architecture, and are trained to convergence. Tables~\ref{tab:abl_distill_design_v2} and~\ref{tab:credit_assignment} report different Cross-teacher values because they come from independently trained ablation families.
All three teachers reach $>98\%$ SR (Table~\ref{tab:abl_distill_design_v2}), indicating the RL framework itself is sound; yet the corresponding students differ sharply: $0\%$ (w/o FOV), $54.8\%$ (w/o C-GV), and $87.3\%$ (ours).
The w/o C-GV student's shorter Time is a success-selection effect: it only succeeds on near-trivial starts, and its Traj.\ is still longer than ours.
Qualitatively, the constrained teacher pushes while facing the anchor inside the gated FOV; the w/o C-GV teacher does not control anchor visibility, and the w/o FOV teacher often pushes facing away. The student gap therefore tracks a behavioral difference between teachers, not a difference in their competence.

\noindent\textbf{Relational Term.}
To isolate the auxiliary term in Eq.~\ref{eq:distill_loss} we distill the same constrained teacher with $\lambda_{\text{rel}}{=}0$, holding the teacher, the online procedure, the interaction budget, and the evaluation layouts fixed. The student drops to $68.7\%$ ($[65.8,71.4]$) on the four-object cross against the $87.3\%$ of Table~\ref{tab:abl_distill_design_v2}, and $74\%$ of its failures are collisions rather than mis-placements. The term constrains the pairwise structure among the active, anchor, and obstacle latents, so what degrades when it is removed is the relation the student needs in order to keep clear of the objects it is not pushing.

\begin{figure}[t]
    \centering
    \begin{minipage}[c]{0.58\columnwidth}
        \centering
        \includegraphics[width=\linewidth]{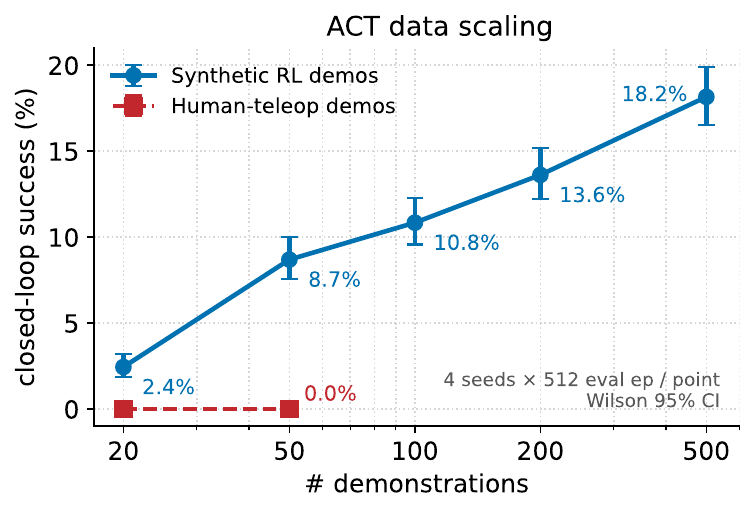}
    \end{minipage}\hfill
    \begin{minipage}[c]{0.40\columnwidth}
        \centering
        \footnotesize
        \setlength{\tabcolsep}{2pt}
        \renewcommand{\arraystretch}{1.05}
        \begin{tabular}{lccc}
            \toprule
            \textbf{Policy} & \textbf{$V$$\uparrow$} & \textbf{$g$$\uparrow$} & \textbf{$\tau$$\downarrow$} \\
            \midrule
            Teacher & 0.91 & +0.62 & 2.1 \\
            \rowcolor{gray!10}
            \textbf{Ours} & 0.90 & +0.61 & 5.9 \\
            ACT~\cite{zhao2023learning} & 0.55 & +0.28 & 50.2 \\
            \bottomrule
        \end{tabular}
    \end{minipage}
    \caption{\textbf{Why ACT saturates.} Left: teleop demos give $0\%$, RL demos saturate at $18.2\%$. Right: active perception over $256$ two-object episodes; $V$ in-view rate, $g$ active gain against a position-matched random-heading null, $\tau$ re-acquisition steps.}
    \label{fig:act_diagnosis}
\vspace{-3mm}
\end{figure}

\begin{table*}[!t]
    \centering
    \caption{Real-World Evaluation\tabnote{\emph{SR} is trial-level: the cross completed with no collision, every centroid within $7.5\,$cm of its target and every orientation within $15^\circ$. Pos./Ori.: mean position and orientation error over every object of every trial; latency: per-frame mean$\pm$std.}}
    \label{tab:realworld}
    \footnotesize
    \setlength{\tabcolsep}{5pt}
    \begin{tabular}{lccccccc}
        \toprule
        Segmenter & \#Tr. & SR & Pos.\,(cm) & Ori.\,(\textdegree) & Img.\,Proc.\,(ms) & Net Infer.\,(ms) & Comm.\,(ms) \\
        \midrule
        HSV (controlled)    & 30 & 21/30 & 4.0 & 5.8 & 26.65$\pm$2.07 & 5.97$\pm$0.88 & 4.46$\pm$0.97 \\
        MobileSAM (cluttered) & 30 & 20/30 & 4.1 & 5.8 & 52.95$\pm$8.33 & 6.19$\pm$1.13 & 5.77$\pm$0.53 \\
        \bottomrule
    \end{tabular}
\end{table*}

\begin{figure*}[!t]
  \centering
  \includegraphics[width=\textwidth]{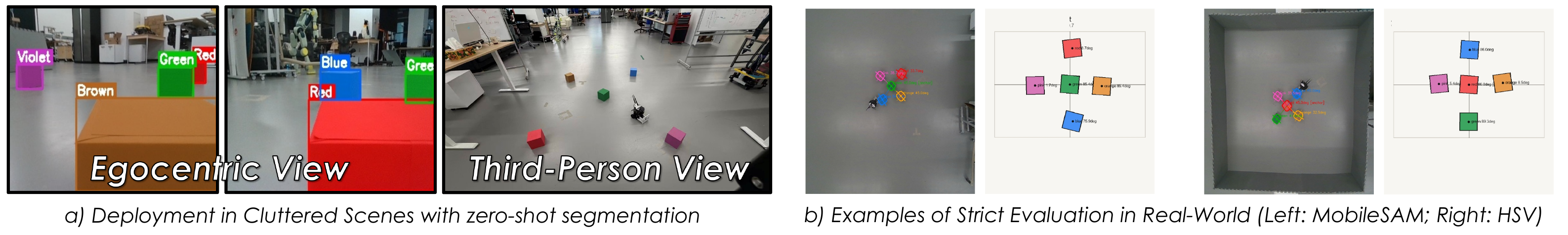}
  \caption{\textbf{Real-world deployment.} (a) MobileSAM-refined segmentation in cluttered scenes (RGB only forms masks). (b) Pose-criterion evaluation rollouts (real vs.\ target layout).}
  \label{fig:realworld_clutter}
\vspace{-5mm}
\end{figure*}

\noindent\textbf{Credit Assignment Ablation.}
We compare \textbf{Base} (episode-terminal success only), \textbf{SWR} (stage-wise rewards from Eq.~\ref{eq:r_reach}/\ref{eq:r_place} without $\eta_t$), \textbf{SWR+TD} (decay on a global episode timer), and \textbf{Ours} (decay reset at each stage boundary). Only this ablation extends the episode limit to $10\,T_s{=}10{,}000$ steps, to make \textbf{Base} measurable at all, which sets the scale in Fig.~\ref{fig:credit_assignment}. \textbf{Ours} stabilizes episode length earliest, at $\sim$45k vectorized steps, while the ablations remain long or oscillatory; reward curves use variant-specific scales. Table~\ref{tab:credit_assignment} compares \textbf{Ours} stopped at convergence against ablations trained to $\sim$90k, twice the budget, an asymmetry that favors them.
Sparse terminal feedback leaves \textbf{Base} unreliable ($16.02\%$ SR). \textbf{SWR} gives the largest single gain, $87.50\%$; the episode timer carries SR to $97.95\%$, and the stage-reset timer to $98.63\%$ with the best Time and Traj.\ of the four. That last difference is within evaluation noise, so the stage timer buys not higher final success but convergence on half the training steps and shorter successful trajectories.

\subsection{Full-System Validation}
\label{sec:validation}

\noindent\textbf{Stage Inference Without Privileged Signals.}
The student's self-maintained stage counter (Sec.~\ref{subsec:phase2}) agrees with the hidden stage index on $84.2\%$ of steps ($256$ episodes, $126{,}784$ steps; mean $|\Delta s|{=}0.52$). The disagreement is the four-step detection latency at each switch plus drift within failed episodes, and its cost is already contained in every reported success rate.

\noindent\textbf{A Second Formation and Unseen Shapes.}
The same pipeline, retrained per task category with nothing else changed, also produces a \textsc{Line-Shape} student. That formation chains its slots: each target is anchored to the previously placed object rather than to one fixed anchor, so a placement error carries into every stage after it. The student reaches $61.3\%$ ($[58.3,64.3]$) against the $87.3\%$ of the cross. Chaining the anchors costs roughly a quarter of the episodes, and the teacher construction and the distillation procedure carry over to it unchanged. Every policy is trained on cuboid active objects; substituting a cylinder or a prism at test time changes the contact dynamics with no adaptation. The student still reaches the object on almost every episode ($99.4\%$/$100\%$), while placement, which does depend on how the object answers a push, completes $67.5\%$/$54.3\%$ (Fig.~\ref{fig:framework}, \emph{Sim Results}). The skill carries to unseen geometries at a cost rather than a collapse.

\subsection{Active Perception Analysis}
\label{sec:active_perception}

We quantify the behavior encouraged by the visibility design using the \emph{active gain} $g$, the in-view rate $V$ minus a
position-matched random-heading null. On the two-object setup (Fig.~\ref{fig:act_diagnosis}, right),
teacher and student have nearly identical gaze ($V=0.91$ vs.\ $0.90$), consistent with transfer of
closed-loop look-to-act behavior, whereas ACT~(RL,\,500) retains only about half the gain ($g=0.28$), a mechanistic complement to Fig.~\ref{fig:act_why}. The same diagnostic orders the observation ablation (Tab.~\ref{tab:abl_distill_design_v2}, student rows). Removing the center gate drops the gain from $+0.59$ to $+0.36$ and SR from $87.3\%$ to $54.8\%$; removing the FOV mask as well drives the gain negative ($-0.19$) and SR to zero.

\subsection{Real World Experiments}
\label{sec:real_world}
We deploy the same student zero-shot in two setups, 30 trials each, on cross-shape rearrangement with a TurtleBot3 Burger, an Intel RealSense D435i, and five $15\,$cm colored cubes: one anchor and four boxes pushed in sequence, a 2-minute timeout, no fine-tuning, and an onboard Jetson Nano streaming RGB-D to a laptop that runs the policy and returns velocity commands. All 60 trials run autonomously to termination with no human input.

The first setup is a controlled arena (Fig.~\ref{fig:head}, left) enclosed by four identical, untextured walls of a single color, segmented by \textbf{HSV} thresholding. The bare walls leave no stable background feature to localize against, so the result cannot be explained by implicit localization off the surroundings.

The second setup adds everyday objects as visual distractors, which take no part in the rearrangement and are never contacted, and segments with color-gated \textbf{MobileSAM}-refined masks~\cite{mobile_sam}. Background appearance never reaches the policy, which consumes role-grouped \emph{depth} with roles matched by identity, so clutter here is a segmentation problem rather than a control one.

The policy transfers zero-shot under both front-ends: $41$ of $60$ trials succeed (Table~\ref{tab:realworld}). Across every object of all $60$ trials, position error averages $4.0\,$cm and orientation error $5.8^\circ$. Failures are dominated by yaw: $16$ of the $19$ place every centroid within tolerance and miss only on orientation.
Closed-loop behavior matches simulation, the robot circling to keep the anchor in its C-GV FOV before contact, and the per-frame round-trip stays above the $10\,$Hz control rate under both front-ends (Table~\ref{tab:realworld}); commanded speeds are capped for the servo torque limit, which slows execution without changing the staged behavior.
Residual failures stem from a box pushed out of the anchor's C-GV FOV after contact with no recovery in the budget, and rare oscillation between goal- and path-seeking views (Sec.~\ref{sec:limitation}).

\section{Limitations and Future Work}
\label{sec:limitation}
EgoPush does not attempt open-world rearrangement; the assumptions of Sec.~\ref{subsec:assumptions} bound it to canonical formations and simplified geometries. The visibility constraint is a frustum-and-range approximation without inter-object occlusion modeling, and the C-GV rule is designed for the present anchor-relative goals. Roles, stage order, and target slots are given by the task (Sec.~\ref{subsec:assumptions}) rather than inferred, so recovering from a wrong one is outside what the system is asked to do; inferring them from an instruction or a demonstration is a separate problem.

One question stays open: whether memory substitutes for the constraint rather than complementing it. The comparison that settles it distills a recurrent student from both the constrained and the unconstrained teacher, with everything else held at the values used here. Only indirect readings are available today: the \textbf{w/o FOV} student's gain is negative ($g{=}{-}0.19$, Tab.~\ref{tab:abl_distill_design_v2}), so its trajectory rarely contains what a memory would integrate, and the recurrent end-to-end baseline reaches $0.10\%$ SR (Tab.~\ref{tab:baseline_comparisons}).

Two limits point at future work. The zero-action stage inference motivates a learned stage or belief state, and so does the student's lack of a belief over unobserved objects, which produces oscillation in two ways: when the goal cue and a free path lie in different directions, turning to one loses the other; and when the placed objects leave the view, two stages present near-identical role-grouped depth while calling for different slots. A belief would address both.

\section*{ACKNOWLEDGMENT}
We would like to thank Eugene Vinitsky, Tianyue Wu, and Tianyu Zhao for valuable discussions. We also thank Zeyu Jiang for assistance with the real-world experiments, Zhejun Cui and Zhichao Han for valuable feedback on early drafts of the manuscript, and Bowen Hu and Weijie Kong for help with figure preparation. This work was supported by NSF Grant 2238968, and by NYU IT High Performance Computing resources, services, and staff expertise.

\bibliographystyle{IEEEtran}
\bibliography{references}

\end{document}